\documentclass[lettersize,journal]{IEEEtran}
\usepackage{graphicx}
\usepackage{cite}
\usepackage{picinpar}
\usepackage{amsmath}
\usepackage{url}
\usepackage{flushend}
\usepackage[latin1]{inputenc}
\usepackage{colortbl}
\usepackage{soul}
\usepackage{multirow}
\usepackage{pifont}
\usepackage{color}
\usepackage{alltt}
\usepackage[hidelinks]{hyperref}
\usepackage{enumerate}
\usepackage{breakurl}
\usepackage{epstopdf}
\usepackage{pbox}
\usepackage{amsmath,amssymb,amsfonts}
\usepackage{algorithmic}
\usepackage{algorithm}
\usepackage{textcomp}
\usepackage{xcolor}
\usepackage{colortbl}
\usepackage{colortbl} 
\usepackage[normalem]{ulem}
\usepackage{booktabs}
\usepackage{threeparttable}
\usepackage{changepage}
\usepackage{makecell}
\usepackage{tabularray}
\usepackage{cuted}
\usepackage[normalem]{ulem}
\usepackage{caption}
\usepackage[caption=false,font=normalsize,labelfont=sf,textfont=sf]{subfig}
\begin{document}

\title{MM-LINS: a Multi-Map LiDAR-Inertial System for Over-Degenerate Environments}

\author{		\vskip 1em
	Yongxin Ma$^{*}$, Jie Xu$^{*}$, Shenghai Yuan$^{\dag}$, Tian Zhi, Wenlu Yu, Jun Zhou$^{\dag}$, and Lihua Xie, \emph{Fellow, IEEE} 
	\thanks{$^{*}$ denotes equal contribution, $^{\dag}$ denotes corresponding author.}
	\thanks{Yongxin Ma and Jun Zhou are with school of Mechanical Engineering, Shandong University, Jinan 250061, China and Key Laboratory of High Efficiency and Clean Mechanical Manufacture, Ministry of Education, Jinan 250061, China (e-mail: yxma@mail.sdu.edu.cn; zhoujun@sdu.edu.cn)}
	\thanks{Jie Xu and Wenlu Yu are with State Key Laboratory of Robotics and Systems, Harbin Institute of Technology, Harbin 150001, China (e-mail: n2308716l@e.ntu.edu.sg; yuwenlu@stu.hit.edu.cn).}
	\thanks{Tian Zhi is with the College of Mechanical and Electrical Engineering, Nanjing University of Aeronautics and Astronautics, Nanjing 210016, Jiangsu, China (e-mail: zt1117@nuaa.edu.cn).}
	\thanks{Jie Xu, Shenghai Yuan, and Lihua Xie are with the School of Electrical and Electronic Engineering, Nanyang Technological University, 639798, Singapore (e-mail: shyuan@ntu.edu.sg; elhxie@ntu.edu.sg).}
	\thanks{\textsuperscript{1} https://github.com/lian-yue0515/MM-LINS}
}



\maketitle

\begin{strip}
\begin{minipage}{\textwidth}\centering
\vspace{-120pt}
\includegraphics[width=1\textwidth]{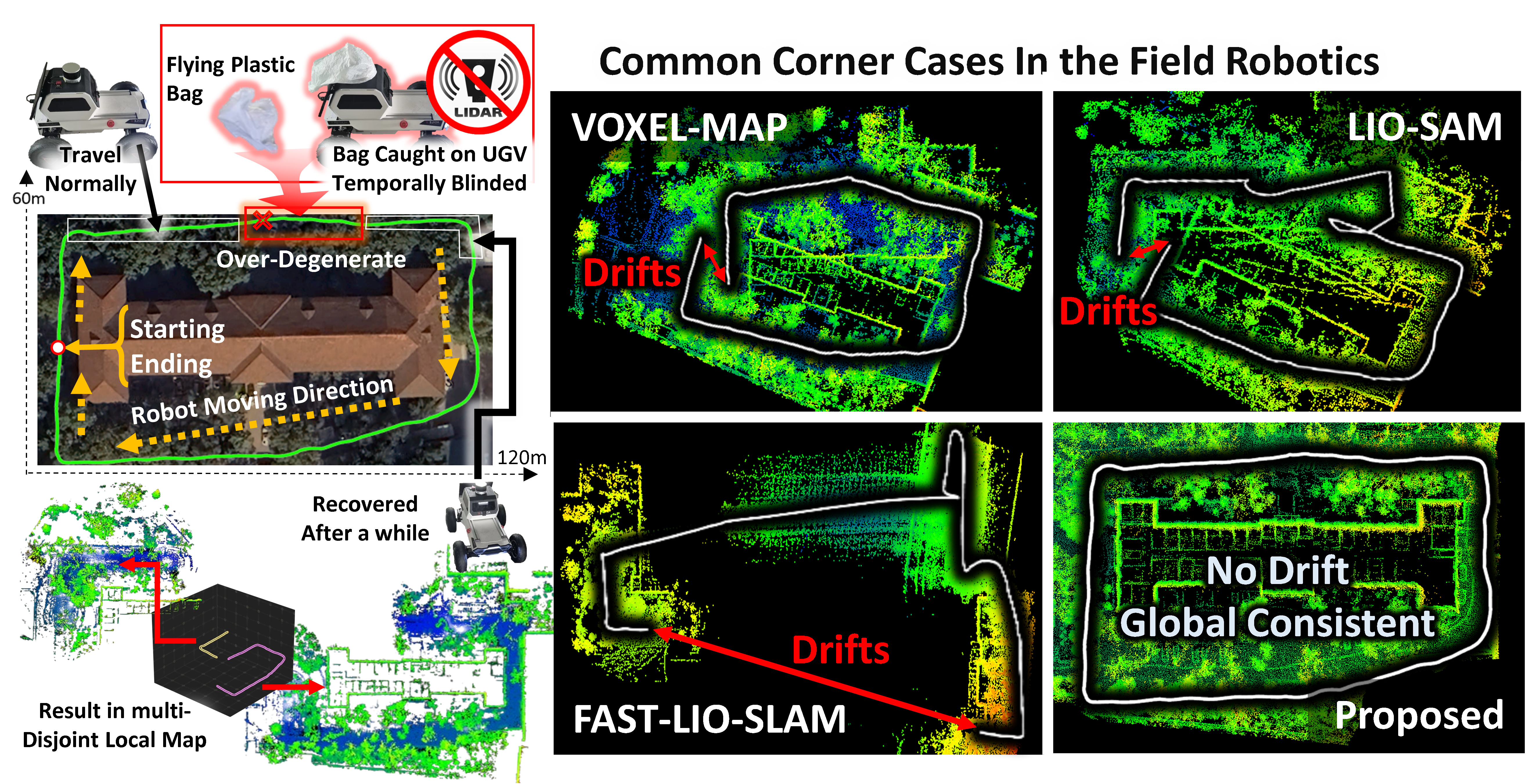}
\captionof{figure}{Left Column: Satellite map and trajectory of the robot during the real-world campus dataset. LiDAR obstruction caused by floating garbage bags is illustrated through real images. Demonstration implementation of our algorithm. 
Right Column: Point cloud maps and trajectories created using SOTA algorithms and ours.}
\label{fig_1}
\end{minipage}
\end{strip}

\begin{abstract}
	SLAM plays a crucial role in automation tasks, such as warehouse logistics, healthcare robotics, and restaurant delivery. These scenes come with various challenges, including navigating around crowds of people, dealing with flying plastic bags that can temporarily blind sensors, and addressing reduced LiDAR density caused by cooking smoke. Such scenarios can result in over-degeneracy, causing the map to drift. To address this issue, this paper presents a multi-map LiDAR-inertial system (MM-LINS) for the first time. The front-end employs an iterated error state Kalman filter for state estimation and introduces a reliable evaluation strategy for degeneracy detection. If over-degeneracy is detected, the active map will be stored into sleeping maps. Subsequently, the system continuously attempts to construct new maps using a dynamic initialization method to ensure successful initialization upon leaving the over-degeneracy. Regarding the back-end, the Scan Context descriptor is utilized to detect inter-map similarity. Upon successful recognition of a sleeping map that shares a common region with the active map, the overlapping trajectory region is utilized to constrain the positional transformation near the edge of the prior map. In response to this, a constraint-enhanced map fusion strategy is proposed to achieve high-precision positional and mapping results. Experiments have been conducted separately on both public datasets that exhibited over-degenerate conditions and in real-world environments. These tests demonstrated the effectiveness of MM-LINS in over-degeneracy environment. Our codes are open-sourced on Github \textsuperscript{1}.
\end{abstract}


\section{Introduction}

Simultaneous \IEEEpubidadjcol localization and mapping (SLAM) are the fundamental components for enabling applications like Mars rover exploration, warehouse AMR, healthcare robotics, restaurant delivery robots, construction mapping, etc.\cite {002,li2024hcto,mcdviral2024}. 
Compared to other visual \cite{xu2023airvo,lyu2023vision,chen2024salient,deng2024compact,deng2024incremental,esfahani2020local,esfahani2019deepdsair,10238802} and radar-based methods \cite{zhang2023ntu4dradlm}, LiDAR SLAM \cite{ji2024lio,nguyen2023slict} can perceive 3D environmental data accurately and in a dense manner \cite{yuan2021survey}, making it a reliable tool for various applications \cite{01,001,003}. 

Existing LiDAR-based SLAM \cite{nguyen2021liro,nguyen2021miliom} struggle in dynamic and cluttered environments \cite{10255288,nielsen2022multi}, such as navigating through crowds of people \cite{ji2022robust}, getting blocked by things like flying plastic bags, or dealing with smoke/fog. These scenarios frequently lead to significant drift issues, which we classify as over-degenerated cases. And there is no simple solution for over-degeneracy.  

Adding proprioceptive sensors like IMUs can partially mitigate the drift issue\cite{1,5,6,nguyen2021viral,nguyen2024eigen}, but prolonged perception interruptions can still affect the overall trajectory and map quality. 
By converting the map representation using different voxel sizes\cite{iG-LIO, 29}, the system can partially address the issue of degeneracy. However, this is not a permanent solution, and the overall map may still become drifted over time.

We hypothesize that in severe cases of over-degeneracy, certain map regions may become irreversibly drifted. This requires reliance on multiple disjoint sub-maps of varying quality to construct a larger map and enable localization. However, this approach poses several challenges, including (1) effectively identifying degeneracy, (2) managing disjoint multi-map sections for determining when and where to use them, and (3) conducting global pose graph optimization with degenerated sections.

This paper presents MM-LINS, a system that focuses on utilizing a multi-map strategy to mitigate the impact of map drift resulting from self-localization failures in over-degenerate areas. It addresses challenges by continuously updating an active map and archiving it into sleeping maps during over-degeneracy. Meanwhile, while in over-degeneracy, the system continuously attempts to construct new maps using a dynamic initialization method to ensure a high-quality initialization of the new map upon leaving these areas. By using Scan Context\cite{8} descriptors, similar keyframes are detected and corresponding maps are reactivated. These maps are then merged for pose graph optimization, resulting in a globally accurate trajectory and map. Experimental results show MM-LINS outperforms traditional algorithms in over-degenerate scenarios (Fig. \ref{fig_1}).

The main contributions of this paper are as follows:
\begin{itemize}
	\item A multi-map LiDAR-inertial SLAM system is proposed for the first time, which mitigates the problem of error accumulation or system failure in LiDAR SLAM systems when faced with over-degenerate scenarios.
	\item In front-end, we propose a reliable degeneracy detection strategy, coupled with a dynamic initialization method for LiDAR SLAM. Regarding back-end, a constraint-enhanced map fusion strategy is proposed, aimed at enhancing the accuracy of map merging.
	\item The system performance is evaluated both in degenerated public datasets and in a real-world environment. The results indicate that MM-LINS can withstand various over-degenerate scenarios, demonstrating higher robustness and accuracy than current state-of-the-art (SOTA) methods under over-degenerate scenarios.
\end{itemize} 

	\section{RELATED WORKS}
Our work intersects the following related fields of degeneracy detection, multi-robot and multi-session SLAM system.

\subsection{Degeneracy Detection}
In this paper, perceiving degenerate phenomena is crucial. Zhang et al.\cite{9} proposed a method for detecting degeneracy in optimization problems. They described the degeneracy detection problem as a state estimation issue in least squares optimization, formulating it as a multicriteria problem. In the filter system, Zhen et al.\cite{10} performed degeneracy judgment based on the Singular Value Decomposition (SVD) of the constraint matrix formed by the normal vectors of the LiDAR beams projected onto the environment plane. X-ICP\cite{X-ICP} utilizes the elements of the Jacobian from the derivation of the matrix for a point-to-plane ICP cost function for degeneracy detection, simplifying the formulation and enabling more practical deployment in various environments. Different from the above methods, we focus on the characteristics of the covariance matrix of the system to improve the efficiency of degeneracy detection.

\subsection{Multi-Robot and Multi-Session SLAM}
Multi-robot and multi-session SLAM has garnered considerable attention in contemporary research endeavors\cite{16,17,18,22,23}. Our work pays additional attention to achieving merging between different maps. Dub\'{e} et al.\cite{12} proposed a robust method for inter-robot closed-loop selection for map fusion aimed at solving relative positional transformations among multiple robots. This method leverages the consistency of the closed-loop and data similarity to prevent mismatches. Huang et al.\cite{14} first implemented the Scan Context descriptor in a multi-robot LiDAR SLAM to address the multi-robot problem, thereby enhancing the data transfer efficiency for public area detection. Do et al.\cite{15} suggested an inter-robot closed-loop selection method to augment the robustness of map fusion. The two-by-two consistency strategy of the internal data introduced in the original closed-loop detection aids in averting false positives in public area detection. Kim et al.\cite{20} put forth the LT-mapper, with its key module, the LT-SLAM, designed to handle multi-session SLAM for long-term operations. The core concept involves solving pose transformations with the Iterative Closest Point (ICP) algorithm and further optimizing through the distance to the loop closure factor. Inspired by the above, global descriptors between different maps are the key to inter-map similarity detection. This paper uses the Scan-context descriptor, which is currently recognized as having better robustness. In contrast to the above approaches, this paper introduces a constraint-enhanced strategy to improve the accuracy of map merging.

\begin{figure*}[!t]
	\centering
	\includegraphics[width=7in]{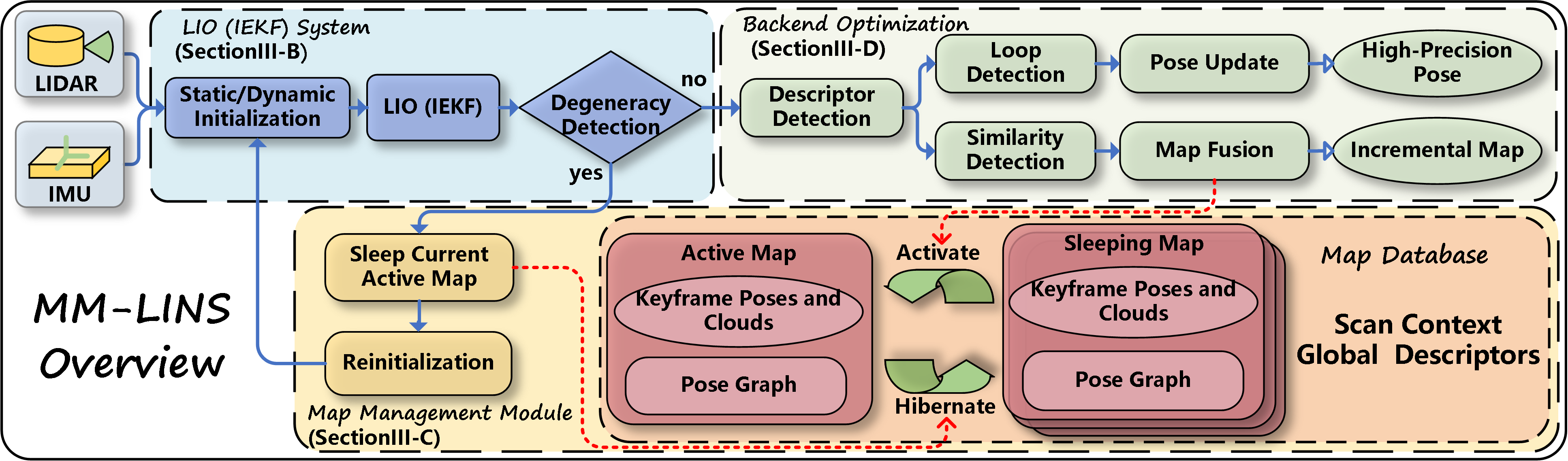}
	\caption{MM-LINS overview.}
	\label{fig_2}
\end{figure*}

\section{METHODOLOGY}
\subsection{Framework Overview}\label{AA}
In this paper, we propose a robust multi-map LiDAR-inertial system framework, dubbed MM-LINS. The proposed framework, shown in Fig. \ref{fig_2}, is designed to operate effectively in environments subject to over-degeneracy. We employs an error state-iterated Kalman filter for pose estimation and propose a new degeneracy detection method to identify over-degenerate scenes. Upon detection of such a scene, the active map is stored into sleeping maps and is reactivated when similar regions are encountered in the future. The system continuously attempts to generate new maps using a dynamic initialization method. Concurrently, the system's back-end leverages feature descriptors based on global geometric information for inter-map similarity detection, enabling the identification of identical regions. When similarities are identified, the corresponding sleeping maps are integrated into the currently active map using an innovative constraint-enhanced map fusion strategy, consequently forming a highly accurate global map upon completion of the system run.

\subsection{Degeneracy-Aware LIO System}\label{B}
The FAST-LIO2\cite{5} is widely recognized as the SOTA LIO system. We have leveraged it as our foundational framework and implemented enhancements to align with the specific demands rising from the challenges. We have innovatively designed and integrated a specialized module for over-degeneracy detection, enabling the system to proactively identify and respond to instances of over-degeneracy. Additionally, a dynamic initialization method has been added to improve performance in new map construction. 

\subsubsection{Over-Degeneracy Detection}
During robot operation, LIO can produce significant errors attributed to LiDAR degeneracy resulting from the lack of geometrically informative structures. This implies that the robot's own 6-DOF are inadequately constrained, particularly concerning the over-degenerate phenomenon described in this paper. This not only leads to the challenge of generating large errors but also poses the risk of system drift or even failure, precisely the issue we aim to proactively mitigate. Hence, The issue we prioritize is on perceiving degenerate phenomena.

For this system, the covariance matrix from \cite{5} contains information about the strength of the constraints for the robot state, which can be divided into submatrices:
\begin{equation}
	\boldsymbol{P}=
	\left[\begin{array}{ll}
		\boldsymbol{P}_{r r} & \boldsymbol{P}_{r t} \\
		\boldsymbol{P}_{t r} & \boldsymbol{P}_{t t}
	\end{array}\right]_{6 \times 6},
\end{equation}
where $r$ represents the rotation, $t$ denotes the translation, and $\boldsymbol{P}_{r r}$ exclusively contains information related to the rotation variables, $\boldsymbol{P}_{t t}$ exclusively contains information related to the translation variables.

It can be explicitly divided into constraints on translations and rotations. Since the scales and types of rotation and translation are different. If analyzed for the covariance matrix as a whole, this would lead to coupling of rotation and translation, resulting in complications with the parameter settings for the degeneracy threshold. Therefore, feature decomposition is performed separately for rotation and translation:
\begin{equation}
	\boldsymbol{P}_{r r}=\boldsymbol{V}_r \Sigma_r \mathbf{V}_r^{\top}, \quad \boldsymbol{P}_{t t}=\boldsymbol{V}_t \Sigma_t \boldsymbol{V}_t^{\top},
\end{equation}
where $\boldsymbol{V}_r$ and $\boldsymbol{V}_t$ are the eigenvectors in matrix, $\Sigma_r$ and $\Sigma_t$ are diagonal matrices with the eigenvalues of $\boldsymbol{P}_{r r}$ and $\boldsymbol{P}_{t t}$ as the diagonal entries, respectively.

The eigenvalues in $\boldsymbol{V}_r$ and $\boldsymbol{V}_t$ provide a direct measure of the strength of the systematic observation with respect to the rotation and translation constraints. Due to the varying magnitudes, separate thresholds are set for their degenerate assessments. As a result, we can monitor the maximum eigenvalues ${\lambda^{(i)}}_{max}^r$ and ${\lambda^{(i)}}_{max}^t$ corresponding to the eigenvalue matrices ${\boldsymbol{V}^{(i)}}_r$ and ${\boldsymbol{V}^{(i)}}_t$ for each frame to evaluate the degeneracy or not of the system.

To achieve this, the major degeneracy threshold ${\xi _{\varkappa}^r}$ and ${\xi _{\varkappa}^t}$, the minor degeneracy threshold ${\xi _{\wp}^r}$ and ${\xi _{\wp}^t}$, and the persistence threshold ${N_{\kappa}}$ of the system are determined empirically. It is worth noting that the relevant threshold settings are closely related to the sensor type. Therefore, we determined the thresholds through experiments on datasets from different sensor types to ensure degeneracy detection performance. These thresholds reflect the occurrence of over-degenerate phenomena and also act as criteria for map hibernation. Whether the over-degeneracy occur is represented in \eqref{eq_2}.
\begin{equation}
	I^{(i)} = \left\{
	\begin{array}{ll}
		1, & \text{if } \lambda^{(i)}_{max}{}^r > \xi_{\varkappa}^r, \lambda^{(i)}_{{max}}{}^t > \xi_{\varkappa}^t \text{ or } \Gamma_\lambda > N_{\kappa} \\
		0, & \text{otherwise}
	\end{array}
	\right. ,
	\label{eq_2}
\end{equation}
where $I^{(i)}$ is the flag for the occurrence of the over-degenerate phenomena in $i_{th}$ frame, $\Gamma_\lambda$ represents the count of consecutive frames with degeneracy falling between the major and minor degeneracy thresholds (${\xi _{\wp}^r} < {\lambda^{(i)}}_{max}^r < {\xi _{\varkappa}^r}$ or ${\xi _{\wp}^t} < {\lambda^{(i)}}_{max}^t < {\xi _{\varkappa}^t}$).

\subsubsection{Initialization Method}
The objective of the initialization module is to estimate pivotal parameters at the onset of the SLAM system: velocity $\mathbf{v}$, gravitational acceleration $\mathbf{g}$, IMU bias $\mathbf{b}_{\mathbf{a}}$, $\mathbf{b}_{\boldsymbol{\omega}}$.

Initially, this system's LIO employs a static initialization method like FAST-LIO2. However, during motion, when our system has experienced over-degeneracy and requires re-initialization for SLAM, the static method fails, making a robust dynamic initialization necessary.  

Extensive research has been conducted in the field of vision for the dynamic initialization of SLAM systems\cite{Visual-Inertia-Initialization, ORB-SLAM2, VINS-Mono}, primarily due to the lack of scale information in monocular systems, necessitating system motion to recover scale. Since LiDAR sensors can directly sense scale information and are robustly initialized in a stationary state, there has been limited focus on their dynamic initialization, which is crucial for our system. Unlike methods that require the estimation of extrinsic calibration of LiDAR and an IMU\cite{Extrinsic_Calibration}, as well as scale information, our approach does not necessitate sufficient excitation in all directions. Inspired by VINS-MONO\cite{VINS-Mono}, our method omits the scale initialization step and directly obtains the LiDAR point clouds with scale, estimating the pose by ICP. Finally, constructs a linear system using the pose and IMU pre-integrations to solve for the initial parameters.

\textit{\romannumeral1) Data Collection:} The gyroscope bias of the parameters to be solved is determined through rotation constraints, while the velocity and gravitational acceleration are resolved via velocity and translation constraints. These constraints are established using at least four frames of data to guarantee solvability of the system. In order to achieve real-time initialization, only the poses computed from four frames of the LiDAR point clouds (i.e., $\boldsymbol{P}_{L_i}^{L}(\boldsymbol{q}_{L_i}^{L}, \boldsymbol{t}_{L_i}^{L}$), where $\boldsymbol{q}_{L_i}^{L}$ and $\boldsymbol{t}_{L_i}^{L}$ represent the ith frame rotation and translation components relative to the first LiDAR frame in the pose, respectively.) and IMU pre-integrations (i.e., $\boldsymbol{\alpha}_{I_{i+1}}^{I_{i}}, \boldsymbol{\beta}_{I_{i+1}}^{I_{i}}, \gamma_{I_{i+1}}^{I_{i}}
$) are collected for subsequent solving.

\textit{\romannumeral2) $\mathbf{b}_{\boldsymbol{\omega}}$ Initialization:} Obtain the rotations $\boldsymbol{q}_{L_{i+1}}^{L}$ and $\boldsymbol{q}_{L_{i}}^{L}$ for two consecutive frames and the relative rotation constraints $\hat{\gamma}_{I_{i+1}}^{I_{i}}$ from IMU pre-integrations. Construct the following equations to solve:
\begin{equation}
	\begin{gathered}
		\min _{\delta \mathbf{b}_{\boldsymbol{\omega}}} \sum_{i=0}^{3}\left\|\mathbf{q}_{I_{i+1}}^{L}{ }^{-1} \otimes \mathbf{q}_{I_{i}}^{L} \otimes \boldsymbol{\gamma}_{I_{i+1}}^{I_{i}}\right\|^2       \\
		\gamma_{I_{i+1}}^{I_{i}} \approx \hat{\gamma}_{I_{i+1}}^{I_{i}} \otimes\left[\begin{array}{c}
			1 \\
			\frac{1}{2} \mathbf{J}_{\mathbf{b}_{\boldsymbol{\omega}}}^\gamma \delta \mathbf{b}_{\boldsymbol{\omega}}
		\end{array}\right],
	\end{gathered}
\end{equation}
where $\mathbf{J}_{\mathbf{b}_{\boldsymbol{\omega}}}^\gamma$ denotes the Jacobi matrix of the relative change in rotation with respect to $\mathbf{b}_{\boldsymbol{\omega}}$, additional note $\boldsymbol{q}_{I_i}^L$ is the same as $\boldsymbol{q}_{L_i}^L$. The solved  $\delta \mathbf{b}_{\boldsymbol{\omega}}$ is the initial  $\mathbf{b}_{\boldsymbol{\omega}}$. 

\textit{\romannumeral3)  $\mathbf{v}$ and $\mathbf{g}$ Initialization:}  Re-propagate IMU pre-integration using $\mathbf{b}_{\boldsymbol{\omega}}$ solved above. A linear measurement model is constructed with relative translation and relative velocity constraints for initial parameter solving. The variables to be solved for this process are as follows:
\begin{equation}
	\boldsymbol{x}=\left[\mathbf{v}_{I_0}^{L}, \mathbf{v}_{I_1}^{L}, \cdots, \mathbf{v}_{I_3}^{L}, \mathbf{g}^{L}\right]^T,
\end{equation}
where $\mathbf{v}_{I_i}^{L}$ is the velocity in the LiDAR coordinates when the $i_{th}$ frame  finishes. $\mathbf{g}^{L}$ is the gravitational acceleration when completes the first LiDAR frame.

Construct the equations as follows:
\begin{equation}
	\begin{gathered}
		\boldsymbol{\alpha}_{I_{i+1}}^{I_{i}}=\mathbf{q}_{L}^{I_i}\left(\mathbf{t}_{I_{i+1}}^{L}-\mathbf{t}_{I_i}^{L}+\frac{1}{2} \mathbf{g}^{L} \Delta t^2 - \mathbf{v}_{I_i}^{L} \Delta t\right) \\
		\boldsymbol{\beta}_{I_{i+1}}^{I_{i}}=\mathbf{q}_{L}^{I_i}\left(\mathbf{q}_{I_{i+1}}^{L} \mathbf{v}_{I_{i+1}}^{L}+\mathbf{g}^{L} \Delta t-\mathbf{q}_{I_i}^{L} \mathbf{v}_{I_i}^{L}\right),
	\end{gathered}
\end{equation}
where $\Delta t$ is the time interval between two frames. Combining the extrinsic calibration of LiDAR and an IMU (i.e., $\boldsymbol{t}_{I_i}^L = \boldsymbol{t}_{L_i}^L - \boldsymbol{q}_{I_i}^L \boldsymbol{t}_{L}^I$, where $\boldsymbol{t}_{L}^I$ denotes the translational extrinsic calibration.) yields the following linear model:
\begin{equation}
	\begin{split}
		\hat{\mathbf{z}}_{I_{i+1}}^{I_{i}} &= \left[\begin{array}{c}
			\hat{\boldsymbol{\alpha}}_{I_{i+1}}^{I_{i}}-\mathbf{q}_{L}^{I_i}\left(\mathbf{t}_{L_{i+1}}^{L}-\mathbf{t}_{L_i}^{L} - \boldsymbol{q}_{I_{i+1}}^L \boldsymbol{t}_{L}^I\right) - \boldsymbol{t}_{L}^I \\
			\hat{\boldsymbol{\beta}}_{I_{i+1}}^{I_{i}}
		\end{array}\right] \\
		&= \mathbf{H}_{I_{i+1}}^{I_{i}} \boldsymbol{x}_{i}+\mathbf{n}_{I_{i+1}}^{I_{i}},
	\end{split}
\end{equation}
where
\begin{equation}
	\begin{gathered}
		\mathbf{H}_{I_{i+1}}^{I_{i}}=\left[\begin{array}{ccc}
			-\mathbf{I} \Delta t & \mathbf{0} & \frac{1}{2} \mathbf{q}_{L}^{I_i} \Delta t^2 \\
			-\mathbf{I} & \mathbf{q}_{L}^{I_i} \mathbf{q}_{I_{i+1}}^{L} & \mathbf{q}_{L}^{I_i} \Delta t
		\end{array}\right], \\
		\boldsymbol{x}_{i}=\left[\mathbf{v}_{I_i}^L, \mathbf{v}_{I_{i+1}}^L, \mathbf{g}^{L}\right]^T,
	\end{gathered}
\end{equation}
solve the linear least squares problem to obtain $\boldsymbol{x}$:
\begin{equation}
	\min _{\boldsymbol{x}_I} \sum_{i=0}^{3}\left\|\hat{\mathbf{z}}_{I_{i+1}}^{I_{i}}-\mathbf{H}_{I_{i+1}}^{I_{i}} \boldsymbol{x}_{i}\right\|^2.
\end{equation}

Next, we refine the gravitational acceleration, as done similarly to \cite{VINS-Mono}, by constraining its magnitude to the corresponding local value. Finally, we rotate all LiDAR frame $(\cdot)^{L}$ to the IMU frame $(\cdot)^{I}$.

\subsection{Map Management Module}
The map management module primarily handles the tasks of storing, converting and merging maps.

\subsubsection{Multi-Map Representation}
The map database, as shown in Fig. \ref{fig_2}, comprises several maps, divided into two categories: an active map that the system updates in real-time and a series of sleeping maps derived from an active map. Each map possesses its own keyframe point clouds, keyframe poses and pose graph. The pose graph is constructed based on the starting node of the active map as an a priori constraint. As the system operates, odometry constraints and loop closure constraints are continuously added to perform pose graph optimization, thereby reducing accumulated errors within the maps. To facilitate information exchange between different maps, the database includes a Scan Context global descriptor, falling into two categories: loop closure detection and similarity detection. Loop closure detection occurs in the active map maintained by the system, while similarity detection takes place in the sequence of sleeping maps. Once similarity detection concludes, the identified similar sleeping maps are activated for map fusion operations.

\begin{figure}[!t]
	\centering
	\includegraphics[width=3.4in]{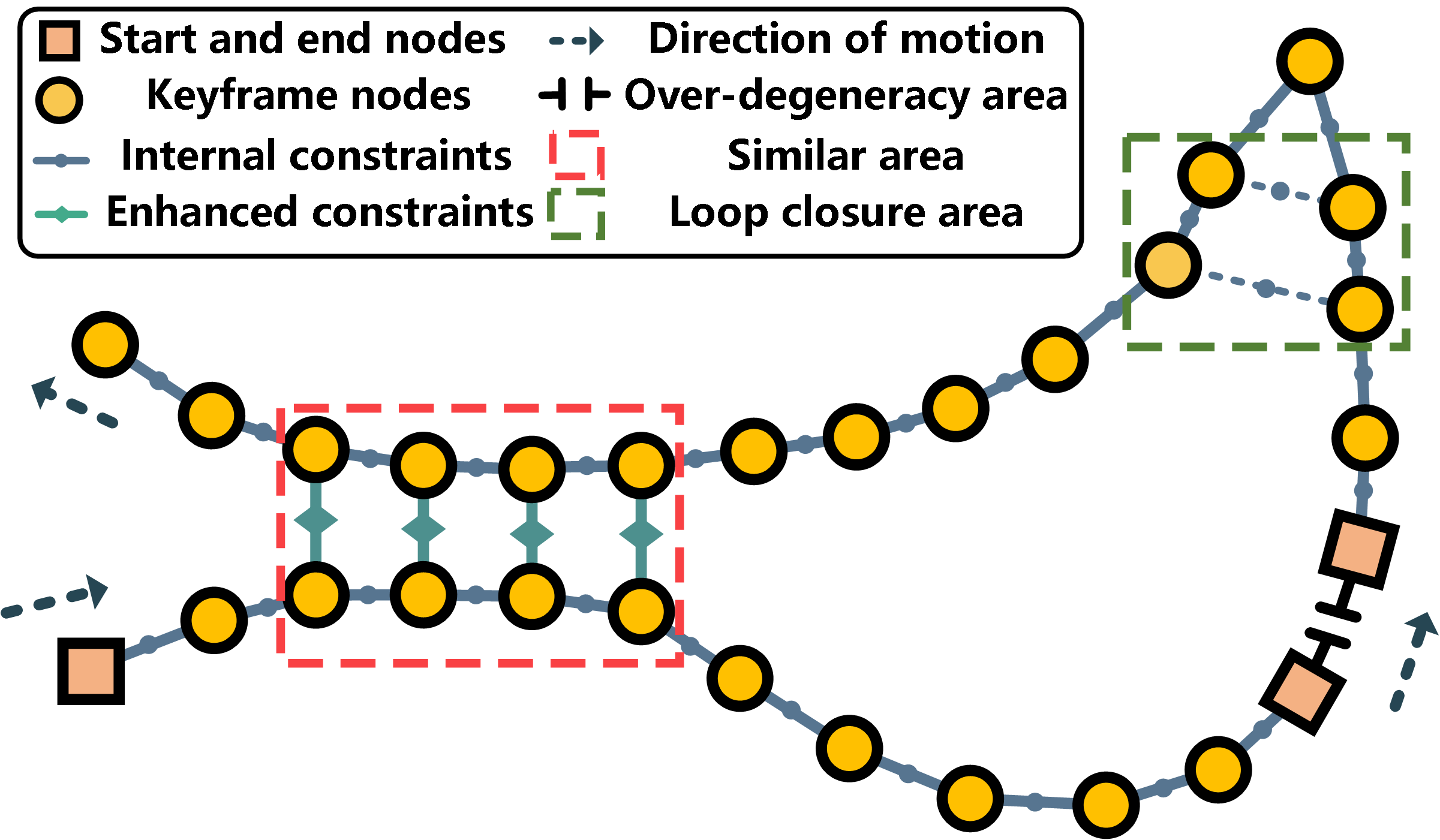}
	\caption{The conceptual overview of the map merging idea. The internal constraints include both loop closure constraints and odometry constraints, which are respectively denoted by dashed and solid lines. Enhanced constraints comprise those detected by similarity measures.}
	\label{fig_3}
\end{figure}

\subsubsection{Map Reconstruction Strategy}
When the over-degeneracy indicator is met, we carry out the ``hibernate" operation to store active map information in sleeping maps. Subsequently, the system continuously employs the dynamic initialization method to ensure new map construction can be completed promptly upon leaving the over-degenerate area.

\subsubsection{Map Fusion}
We employ subscripts $a,$ $s$ and $m$ to distinguish between active, sleeping and merged maps, respectively. First, through similarity detection, we identify keyframe point clouds of the common area, represented by ${C_a}$ and ${C_s}$, between ${M_a}$ and ${M_s}$, and calculate the initial pose transformation, $T_a^s$. We apply $T_a^s$ to all keyframe poses and point clouds in map ${M_a}$ to transform them to the ${M_s}$ coordinate system, and merge ${M_s}$ and ${M_a}$ into ${M_m}$. Subsequently, we carry out pose graph fusion. During this process, the pose graph of ${M_a}$ is aligned with the pose graph of ${M_s}$ in preparation for fusion. However, this transformation renders the a priori constraint in the pose graph of ${M_a}$ unreliable, prompting us to completely discard the priori constraint in the pose graph of ${M_a}$. As a result, the pose graph becomes under-defined. To counteract this issue, we introduce similarity constraints, which in turn ensure the completeness of the pose graph. Lastly, the system carries out pose graph optimization to finalize the optimization of $T_a^s$, and ${M_s}$ is used as the subsequent ${M_a}$ for updates.

\subsection{Constraint-Enhanced Pose Graph Optimization}\label{D}
The back-end primarily focuses on accurate pose estimation and global consistent point cloud mapping. Accordingly, the back-end takes in the odometry poses and point clouds output from the front-end and further optimizes them utilizing the pose graph. Building on this, the back-end framework concentrates on stitching together the multi-maps derived after experiencing over-degeneracy to create a high-precision global map. The conceptual overview of the map merging idea is presented in Fig. \ref{fig_3}.

\begin{table*}[!t]
	\centering
	\caption{Comparison of ATE (RMSE) in meter Across Various Degenerate Datasets}
	\resizebox{\textwidth}{!}{%
		\begin{tabular}{lccccccc}
			\toprule
			\multirow{2}{*}{\makecell{\textbf{Sequences}  \textbf{(Range:m)}}} &  \multirow{2}{*}{\textbf{FAST-LIO2}} & \multirow{2}{*}{\textbf{VOXEL-MAP}} & \multirow{2}{*}{\textbf{LIO-SAM}} & \multirow{2}{*}{\textbf{FAST-LIO-SLAM}} & \multirow{2}{*}{\textbf{FAST-LIO-SAM}} & \multicolumn{2}{c}{\textbf{MM-LINS}} \\ 
			\cmidrule(lr){7-8}
			& & & & & & \textbf{w/o} & \textbf{PGO} \\
			\midrule
			M2DGR-07-31 (289.7m) & 23.65 & 13.27 & 1049.781 & 8.653 & 5.221  & \underline{0.346} & \textbf{0.322} \\
			M2DGR-08-06 (340.6m) & 35.951 & 41.959 & 1177.675 &  25.814 & 389.838 & \underline{0.221}  & \textbf{0.186} \\
			UrbanLoco-04-26 (741.3m) & 38.667 & 26.472 &{$\times$\textsuperscript{a}} & 76.571 & 28.292 & \underline{1.979} & \textbf{1.978} \\
			UrbanLoco-03-16 (601.1m) & 7.27 & 36.697 & {$\times$} & 10.549 & 14.109 & \underline{1.157} & \textbf{1.155} \\
			NCLT-01-10 (1139.2m) & 0.997 &{-- \textsuperscript{b}} & {--}  & 29.146  & 1.203  & \textbf{0.785} & \underline{0.811} \\
			UTBM-07-16 (5044.9m) & 25.718 & 25.481 & 720.591 & 81.736  & 17.817  & \underline{5.118}  & \textbf{2.362} \\
			UTBM-04-18 (5112.3m) & 67.481  & 95.458 & 1845.723  & 96.728 & 43.345 & \underline{5.945} & \textbf{5.033} \\ 
			\bottomrule
		\end{tabular}
	}
	\vspace{1pt}
	\begin{adjustwidth}{0cm}{0cm}
		\footnotesize{\textsuperscript{a} ``$\times$'' denotes the result drifted, and the table below follows the same rules. }\\
		\footnotesize{\textsuperscript{b}  ``--'' denotes numerical instability leading to crashing either at Start-up or at degeneracy encounter. }\\
		\footnotesize{  Best results are \textbf{boldened}, and second-best results are \underline{underlined}, and the table below follows the same rules. }
	\end{adjustwidth}
	\label{tab1}
\end{table*}

\begin{figure*}[!t]
	\centering
	\includegraphics[width=5.5in]{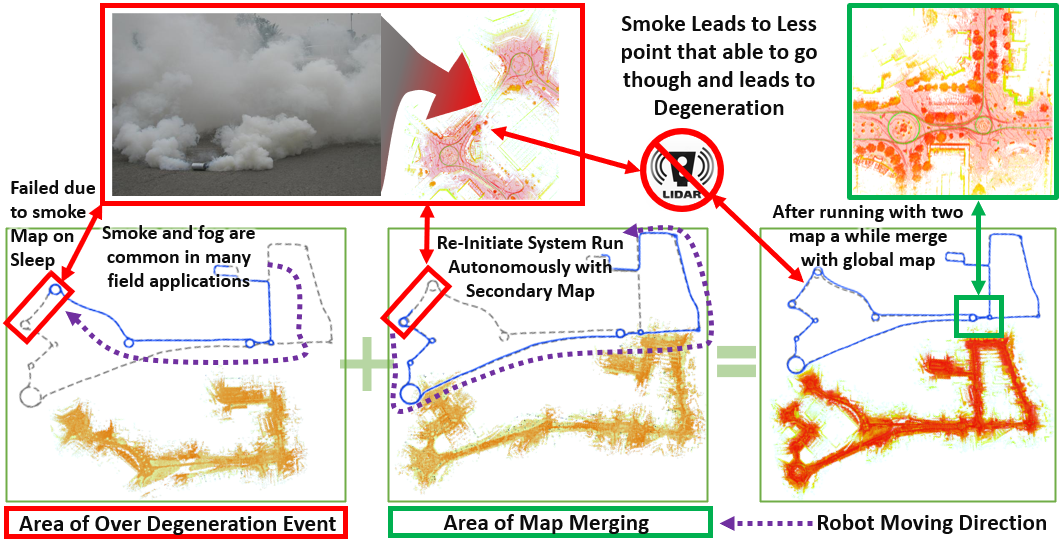}
	\caption{A visual demonstration of degeneracy event and map merging.}
	\label{fig_4}
\end{figure*}

\subsubsection{Descriptor Detection}
This system employs the Scan Context descriptor\cite{8} for loop closure detection and similarity detection. To meet the demands of map merging, we impose stringent screening conditions for similarity detection to avoid misidentification of similar maps. We hypothesize that similarity detection does not usually occur by chance in a single frame, but occurs continuously. Therefore, we augment the original loop closure detection with a time consistency condition for similarity detection.

The time consistency check essentially quantifies the number of consecutive occurrences of similarity detection (as shown in \eqref{eq_3}). A similarity detection is only accepted if the similarity consistently occurs over a successive period. This can influence the quality of the subsequent optimization for the map pose transformation.
\begin{equation}
	\Gamma_{\varsigma} > {\varepsilon _{th}},
	\label{eq_3}
\end{equation}
where $\Gamma_{\varsigma}$ signifies the count of similar Scan Context features occurring consecutively, while ${\varepsilon _{th}}$ represents the threshold of temporal consistency, a value determined based on the experimental environment.

\subsubsection{Constraint-Enhanced Fusion Optimization}
The back-end maintains the active map in real-time by building a keyframe pose graph, where a new keyframe is initiated when the distance or angle between adjacent frames exceeds a certain threshold. As the robot moves, optimization is performed by adding odometry constraints and loop closure constraints. In this context, Let $P = \left\{ {p_0}, \ldots ,{p_t}\right\}, p \in SE(3)$ represent a set of keyframe poses filtered from the odometry information passed from the front-end, encompassing a set of 6-DoF robot poses from frame $0$ to frame $t$. $C$ encompasses all constraints within the map, including both the odometry constraints inferred between keyframe nodes and the loop closure constraints identified through loop detection. For each pair of pose constraints, denoted as $\left( {i,j} \right) \in C$, we define the error between the observed transformation ${O_{ij}} \in SE(3)$, corresponding to the loop constraint between nodes, and the predicted transformation ${\widehat O_{ij}} \in SE(3)$, derived from odometry inference, as ${e_{ij}}$ :
\begin{equation}
	{e_{ij}}\left( {{p_i},{p_j},{O_{ij}}} \right) = {O_{ij}}\boxminus{\widehat O_{ij}}\left( {{p_i},{p_j}} \right),
	\label{eq_4}
\end{equation}
where $\boxminus$ encapsulation operator denotes a mapping from a local neighborhood on $SE(3)$ to its tangent space\cite{4}, and ${\widehat O_{ij}}\left( {{p_i},{p_j}} \right) = p_i^{-1}{p_j}$. 

This optimization problem can be represented as a nonlinear least squares problem in \eqref{eq_5}, The system aims to optimize the robot's pose to minimize the error.

\begin{equation}
	{P^*} = \mathop {\arg \min }\limits_P \sum\limits_{(i,j) \in C} {{R_{ij}}},
	\label{eq_5}
\end{equation}
\begin{equation}
	{R_{ij}} = e_{ij}^T{\Omega _{ij}}{e_{ij}},
	\label{eq_6}
\end{equation}
where ${R_{ij}}$ signifies the negative log-likelihood function of a constraint between ${p_i}$ and ${p_j}$, ${\Omega _{ij}}$ stands for the covariance.

When map fusion is not occurring, the back-end optimization process is shown by the above. During map merging, the primary expectation is to accurately solve the pose transformation between the maps, which directly affects the integrity and consistency of the global map. Therefore, the map optimization objective changes, as described below.

In the context of merging two maps, the corresponding subindexes are defined as follows: $\psi  = \{ a,s\}$, $a$, and $s$ represent the active and sleeping maps, respectively. $\forall \alpha  \in \psi$, ${P_\alpha }$ denote the set of poses for map $\alpha$ . $ \mathbb{P} = \left\{ {{P_\alpha }\mid \alpha  \in \psi } \right\} $ contains all poses of the map being merged. ${C_\alpha}$ represents all pose constraints within $M_a$ and $M_s$, and ${C_a^s}$ denotes the inter-map transformation constraints between $M_a$ and $M_s$. With these definitions, the variables to be optimized are as follows:
\begin{equation}
	\mathbb{X} = \left\{\mathbb{P}, {T_a^s} \right\}.
	\label{eq_7}
\end{equation}

For the optimization problem, the equation is as follows:
\begin{equation}
	\mathbb{P}^* = \mathop {\arg \min }\limits_P \left\{\sum_{(i,j) \in C_\alpha}{R_{ij}(\mathbb{X}) + R_a^s(\mathbb{X})} \right\},
	\label{eq_8}
\end{equation}
where is composed of two components: The first half consists of the negative log-likelihood function of constraints in intra-map and the second half is the negative log-likelihood function of inter-map transformation.

Due to the inability to provide an accurate constraint relationship between the maps directly, we constrain the transformation relationship between the maps by fixing the overlapping part of the trajectory of the active map and the sleeping map in the common area. If only one similarity constraint occurs during merging, the accuracy of map merging is not high enough. Therefore, a constraint-enhanced strategy is proposed, as shown in Fig. \ref{fig_3}, during map merging, we optimize the pose transformation between maps using a certain number of similarity constraints to form enhanced constraints. The objective function is modified into the following form:
\begin{equation}
	\mathbb{P}^* = \mathop {\arg \min }\limits_P \left\{\sum_{(i,j) \in C_\alpha}{R_{ij}(\mathbb{X}) + \sum_{C_{\Upsilon}}R_{\iota}(\mathbb{X})} \right\},
	\label{eq_9}
\end{equation}
where $R_{\iota}(\mathbb{X})$ denotes the enhanced constraints between maps, composed of pairs of similar constraints between two maps, indicated by $\mathop {\left( {i,j} \right)}_{i \in \alpha, j \in \beta, \alpha \ne \beta}  \in {C_{ \Upsilon}}$. It is evident that a higher quantity of these similarity constraints (indicating more pronounced constraint enhancement) leads to more accurate calculations of pose transformations. This quantity is determined by the threshold ${\varepsilon _{th}}$ in the similarity detection.

\section{EXPERIMENT RESULTS}
In this work, we extensively evaluated the efficacy of our proposed MM-LINS system using both public datasets and real-world field-collected proprietary datasets. All experimental procedures were executed on an old laptop equipped with an Intel Core i7-9750H (2.60GHz) CPU.

\begin{table}[!t]
	\centering
	\caption{End-to-End Distance Comparison (meter) of Real-World Proprietary Datasets (Indoor)}
	\begin{tabular}{cccccccc}
		\toprule
		\textbf{Sequence } & \textbf{Fast} & \textbf{Voxel} & \textbf{Fast-LIO} & \textbf{Fast-LIO} & \multicolumn{2}{c}{\textbf{MM-LINS}} \\ 
		\cmidrule{6-7}
		\textbf{(Range:m)} & \textbf{LIO2} & \textbf{Map} & \textbf{SLAM} & \textbf{SAM} & \textbf{w/o} & \textbf{PGO}  \\ 
		\midrule
		Fac.1(87.2)  & 29.6 & 67.26 & 116.57 & $\times$ & {\underline{0.22}} & {\textbf{0.18}}  \\
		Fac.2(88.1)  & 27.25    & 23.01    & 66.60  & \underline{0.11} & 0.15 & \textbf{0.05} \\
		Fac.3(96.9)   & 1.41     & 8.81     & 25.49        & 17.35                             & {\underline{0.53}} & {\textbf{0.31}} \\
		\bottomrule
	\end{tabular}
	\vspace{1pt}
	\begin{adjustwidth}{0cm}{0cm}
		\footnotesize{ ``Fac.'' denotes handheld factory dataset with a single event of Over-Degeneracy.}
	\end{adjustwidth}
	\label{tab2}
\end{table}

\begin{figure*}[!t]
	\centering
	\includegraphics[width=4.5in]{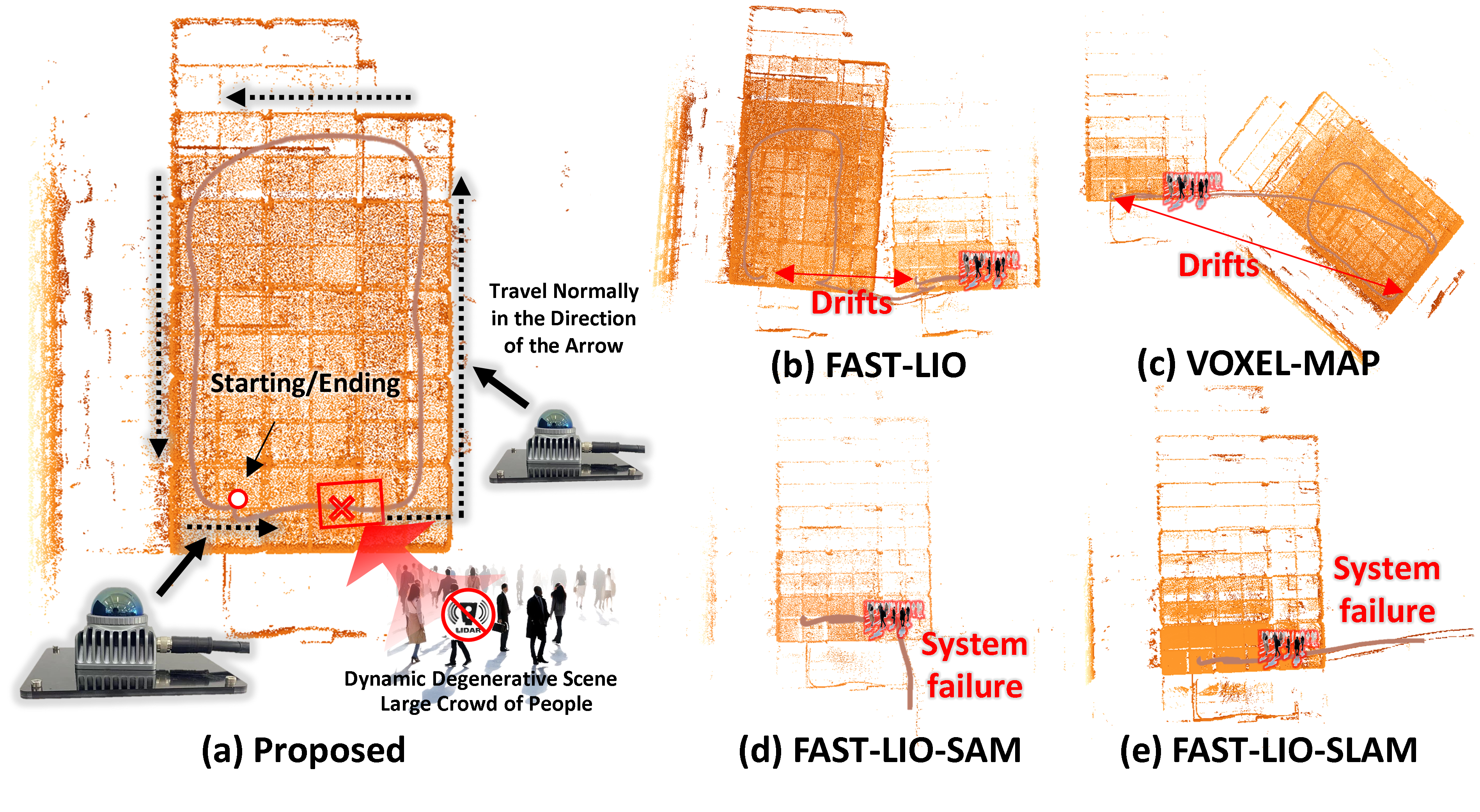}
	\caption{Performance comparison of different algorithms on the Factory-1 dataset. The subfigures depict the following: (a) Real-world operation of the handheld device and the excellent performance of our algorithm in this case, (b) - (e) Point cloud maps and trajectories of SOTA algorithms in the same situation.}
	\label{fig_5}
\end{figure*}

\begin{table*}
	\centering
	\caption{End-to-End Distance Comparison (meter) of Real-World Proprietary Datasets (Outdoor)}
	\begin{tblr}{
			colspec={Q[c,m] *{8}{Q[c,m]}},
			cell{1}{1} = {r=2}{},
			cell{1}{2} = {r=2}{},
			cell{1}{3} = {r=2}{},
			cell{1}{4} = {r=2}{},
			cell{1}{5} = {r=2}{},
			cell{1}{6} = {r=2}{},
			cell{1}{7} = {c=2}{},
			cell{1}{9} = {r=2}{},
			cell{3}{9} = {r=3}{},
			hline{1,3,6-8} = {-}{},
		}
		\textbf{Sequences (Range:m)}          & \textbf{FAST-LIO2}  & \textbf{VOXEL-MAP} &\textbf{ LIO-SAM} & \textbf{FAST-LIO-SLAM}  & \textbf{FAST-LIO-SAM} & \textbf{MM-LINS}          &                & Remark                                         \\ \cline{7-8}
		&            &           &         &                &              & \textbf{w/o}            & \textbf{PGO}            &                                                \\
		Campus-1 (350.6m)  & 29.705     & 13.026    & 19.941  & 16.610         & $\times$            & \uline{0.126}  & \textbf{0.092} & {Single Event. } \\
		Campus-2 (351.1m)  & 391.773    & $\times$         & $\times$       & 62.072         & $\times$            & \textbf{0.603} & \uline{1.295}  &                                                \\
		Campus-3 (347.6m)  & 43.995     & 67.96     & $\times$       & 7.586          & 8.014        & \uline{0.137}  & \textbf{0.093} &                                                \\
		Campus-4 (348.7m)  & 137.555    & 146.381   & 34.152  & $\times$              & 91.339       & \uline{0.128}  & \textbf{0.051} & Two Events.                                     \\
		Campus-5 (349.4m)  & 306.398    & $\times$         & $\times$       & 44.195         & $\times$            & \uline{0.159}  & \textbf{0.132} & Three Events.                                   
	\end{tblr}
	\vspace{2pt}
	\begin{adjustwidth}{0cm}{0cm}
		\footnotesize{All campus series of proprietary datasets are collected via vehicle-mounted perception suits following exactly the same path. The only difference is the physical interruption of sensing caused by wrapping a plastic bag on top of the LiDAR. The number of events denotes the number of physical interferences used to simulate the degeneracy.}
	\end{adjustwidth}
	\label{tab3}
\end{table*}
\begin{figure*}[!t]
	\centering
	\includegraphics[width=7in]{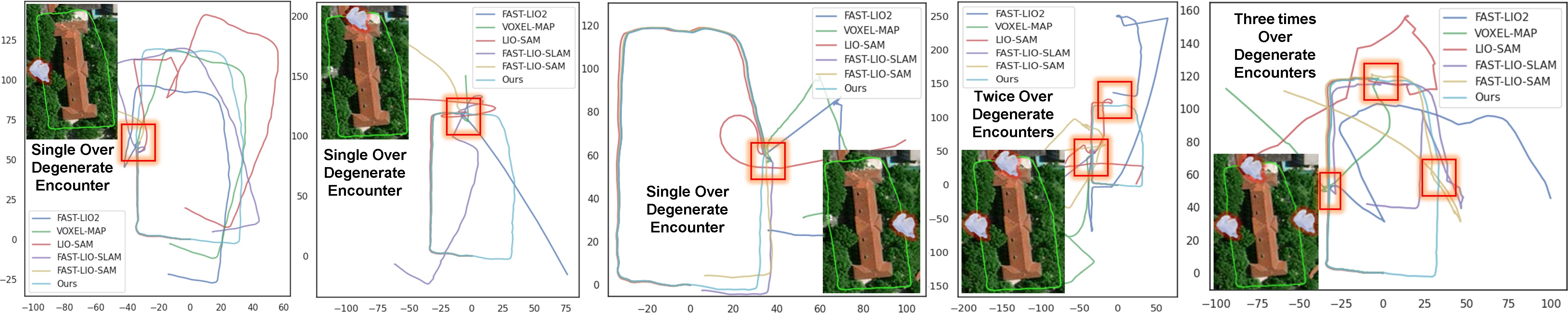}
	\caption{Robot trajectories in campus scenes highlight LiDAR obstruction by drifting garbage bags on various occasions. Compared with various state-of-the-art methods, MM-LINS consistently performs in challenging conditions. }
	\label{fig_7}
\end{figure*}

\subsection{Public Dataset Validation}
We conducted thorough experiments on SOTA methods and MM-LINS using publicly available datasets, with a special focus on scenarios characterized by loop closure. These scenarios assist in identifying maps with analogous regions and facilitate map fusion. Given the lack of real-world public datasets for accurate ground truth, over-degeneracy, and loop closure characteristics, we opted to modify part of the public datasets to simulate over-degeneracy. To streamline the modification process, we significantly downsampled the point clouds for a duration prior to the loop closure phase of the dataset, thereby emulating a period of limited observational information and producing an effect analogous to over-degeneracy. To validate the algorithm, we selected seven sequences from four reputable public datasets: M2DGR\cite{25}, NCLT\cite{26}, utbm\cite{27}, and UrbanLoco\cite{28}. The selected sequences encapsulate a diverse range of scenarios, including large-scale urban environments featuring ramps and high-rise buildings, as well as suburban terrain. Each dataset is equipped with different LiDAR and IMU configurations, showcasing the versatility and applicability of the proposed system across various environments and hardware.

\begin{figure}[t]
		\centering
		\includegraphics[width=1\linewidth]{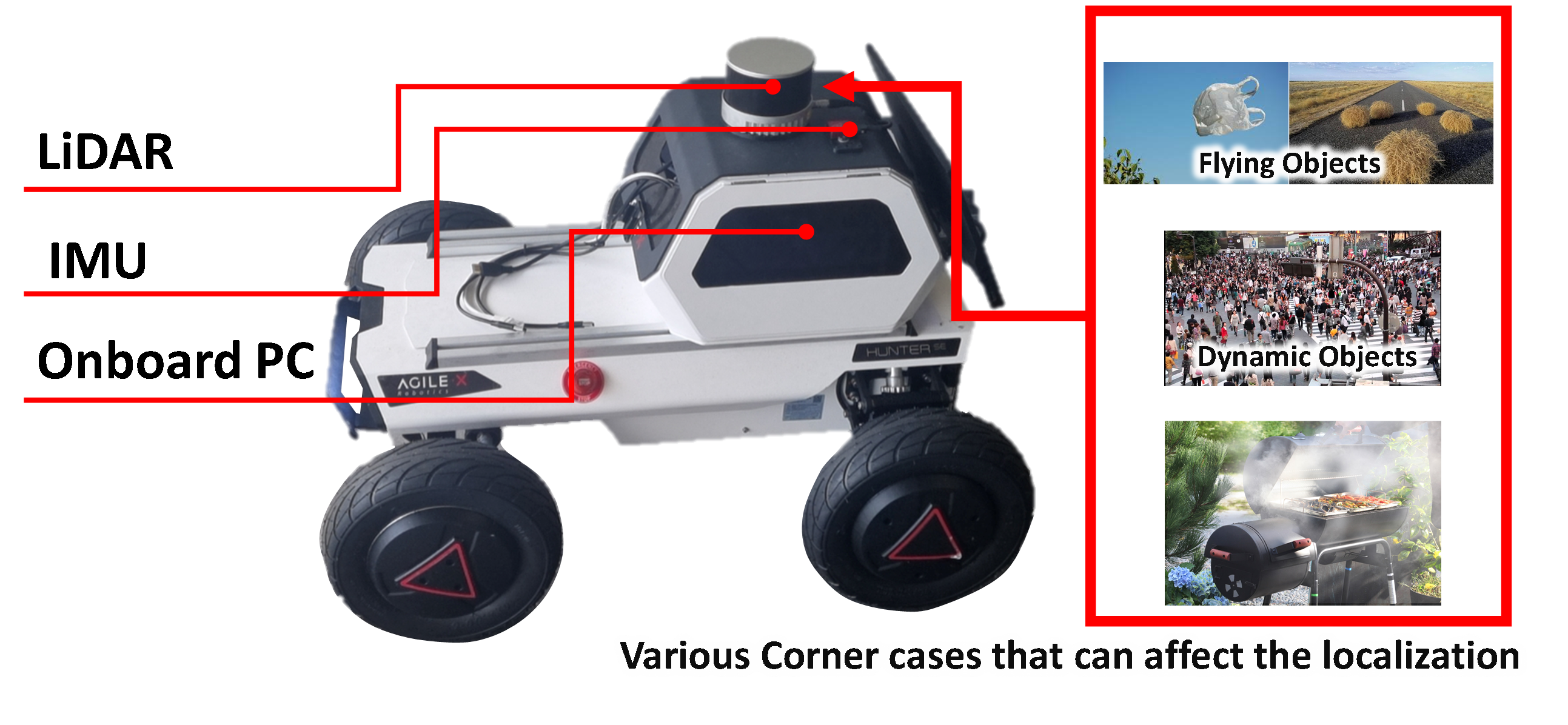}
		\caption{Vehicle-mounted device used for the experiment.}
		\label{fig_6}
		\vspace{-0cm}
\end{figure}

In our multi-map strategy, when it encounters an over-degenerate scenario during mapping, it archives the active map into sleeping maps and initiates the construction of a new map. Upon completion of subsequent similarity detection, map fusion becomes feasible. Fig. \ref{fig_4} provides an example of map fusion. It is apparent that our algorithm can more resiliently navigate the challenges posed by over-degenerate scenarios.

To quantitatively analyse the quality of the mapping, we conducted a comparative analysis of the absolute trajectory errors (ATE) between the trajectories generated by the algorithms and the ground truth. To highlight the accuracy and robustness of our approach, we contrasted the results of FAST-LIO\cite{5}, VOXEL-MAP\cite{29}, LIO-SAM\cite{3}, FAST-LIO-SLAM\cite{7}, and FAST-LIO-SAM\cite{30} algorithms under degeneracy augmented datasets, as delineated in Table \ref{tab1}. The experimental results indicate that the multi-map strategy proposed in this paper generally maintains the error within a range of 5m or lower. Compared to the SOTA methods, it can more effectively manage over-degenerate scenarios and enhances robustness. Ablation studies were further conducted on constraint-enhanced pose graph optimization methods (labeled as \textbf{MM-LINS (w/o PGO)} and \textbf{ MM-LINS (PGO)} in the table). The findings demonstrated a marginal reduction in algorithmic error following the incorporation of the constraint-enhanced method, thereby affirming the necessity of this method.

\subsection{Real-world Field Testing}
It is crucial to ensure that the algorithm works reliably in real-world situations. We tested it out in various environments, both indoors and outdoors, using handheld and vehicle-mounted setups. It is important to highlight that we ensured consistent start and end points for each test, serving as our ground truth measurement. This precaution was particularly necessary because the testing area was densely populated with trees, making RTK/GPS unreliable, Leica Tracking unfeasible and prior map matching inaccurate. By relying on end-to-end distance measurement, we were able to assess the SLAM performance accurately.

\subsubsection{Indoor Experiment}
In this environment, data collection was done by a handheld device equipped with a MID-360 LiDAR. The environment mapped was a factory spanning an area of over 1200 m$^2$. Data collection involved making a complete circuit within the factory premises, during which over-degeneracy was simulated by obstructing the LiDAR while in motion. The specific results are displayed in Table \ref{tab2}. All algorithms, barring the one outlined in this paper, exhibited errors of over 1m, sometimes even higher. This is an unacceptable margin of error for a SLAM system. Conversely, our algorithm was able to constrain the error within 0.4m. Meanwhile, we also show the performance of different algorithms, as shown in Fig. \ref{fig_5}.

\subsubsection{Outdoor Experiment} 
The efficacy of the proposed MM-LINS method was also evaluated outdoors. Data collection was done by a vehicle-mounted device equipped with RS-LIDAR-16 LIDAR and IMU in KY-INS180-A0, as shown in Fig. \ref{fig_6}. Fig. \ref{fig_7} illustrates scenarios of moving a complete circuit around a building on the campus with over-degeneracy occurring at different locations. Additionally, it presents the trajectories of different algorithms in their corresponding scenarios.  The specific results are displayed in Table \ref{tab3}. MM-LINS can handle the case of over-degeneracy at any position as well as a certain number of over-degeneracies, limiting the error to less than 15 centimeters. Other existing algorithms are subject to huge errors unacceptable to the SLAM system when faced with the same situation.

\subsection{Degeneracy Detection and Dynamic Initialization Performance}
\begin{table}[!t]
	\centering
	\caption{The Number of Sub-maps Generated by the System with Different Degeneracy Detection Methods}
	\begin{tabular}{ccccc}
		\toprule
		\textbf{Sequence } & \textbf{Zhang's} & \textbf{Ours} & \makecell{\textbf{N.DE}} & \textbf{LiDAR type}  \\  
		\midrule
		M2DGR-07-31   & 9     & 10     & 9  &\multirow{3}{*}{Velodyne}\\
		M2DGR-08-06   & 11     & 10     & 9  &\\
		NCLT-01-10  & 10 & 10 & 9 &\\
		\bottomrule
	\end{tabular}
	\vspace{1pt}
	\begin{adjustwidth}{0cm}{0cm}
		\footnotesize{ ``N.DE'' denotes the number of degeneracy events.}
		\footnotesize{The theoretical count of submaps is one more than the number of degeneracy events.}
	\end{adjustwidth}
	\label{tab4}
\end{table}

\begin{figure}[t]
	\centering
	\includegraphics[width=1\linewidth]{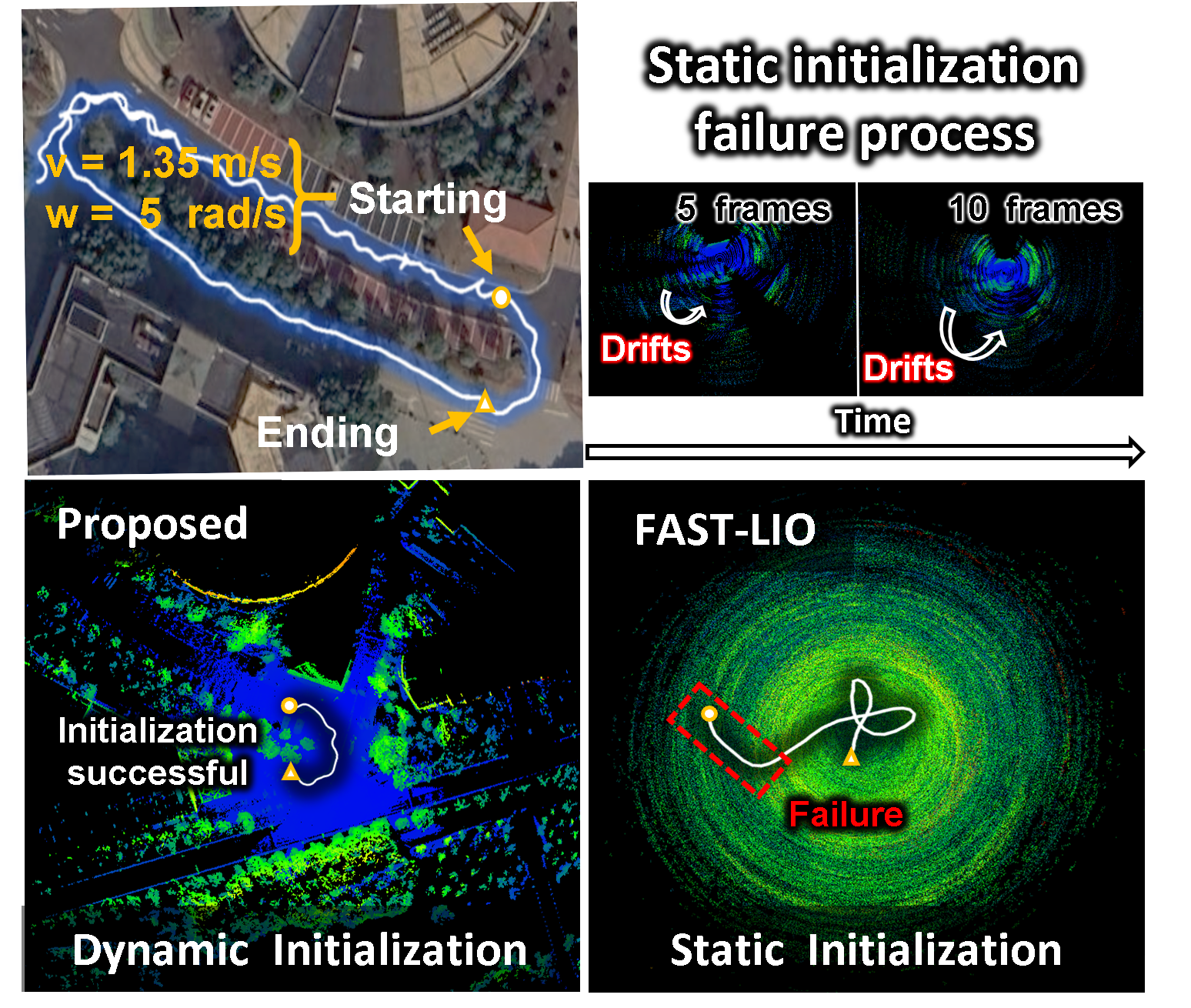}
	\caption{Dynamic initialization performance demonstration on the M2DGR-08-06 sequence. The process begins playing the sequence at the 220s mark (during the robot's traversal) and lasts for 50 seconds. The static initialization method of FAST-LIO exhibits rotational drift, leading to system failure, whereas our method completes initialization and commences mapping.}
	\label{fig_8}
	\vspace{-0.5cm}
\end{figure}

For degeneracy detection, Zhang et al.\cite{9} proposes a degeneracy detection metric called the degeneracy factor, which is considered SOTA. It directly perform singular value decomposition on the information matrix. We used it as a baseline method for comparison in this work, highlighting the superior performance of our method. Given that degeneracy detection is difficult to measure through objective metrics. We applied different degeneracy detection methods to our system with map fusion disabled. The degeneracy detection performance was evaluated by comparing the number of sub-maps generated by the system and the number of degeneracy scenarios. To ensure the number of degeneracy scenarios is controllable, we modified some public datasets to simulate different levels of degeneracy scenarios, while keeping their quantity within a certain limit. Zhang's method sets the threshold to 100, while our method sets the translation and rotation thresholds to 0.005 and 0.0006, respectively. The results are shown in Table. \ref{tab4}. Zhang's method exhibits mismatches in quantity, whereas ours does not. This reflects that separating rotation and translation components can improve detection performance.

For LiDAR-inertial systems, there is a scarcity of research on dynamic initialization. We employed both dynamic initialization and static initialization methods to start map construction as the robot moves. By observing the map quality, we further highlight the importance of dynamic initialization research, as shown in Fig. \ref{fig_8}.

\subsection{Computation Efficiency}
The time consumed in each module is shown in Table. \ref{tab5}. For the NCLT-01-10, UTBM-04-18, and Campus-1 sequences, degeneracy scenarios occurred only once (DI and CE-PGO executed only once), with all modules consuming less than 1.5\% of the total time, which indicates a low additional computation cost compared to map construction. Overall, ODD accompanies the system throughout its lifetime but consumes minimal time. DI's time consumption is mainly related to the number of submaps, CE-PGO's time consumption is related to both the scale and the number of submaps. The larger the number and scale of submaps, the more time they consumes.
\begin{table}[!t]
	\centering
	\caption{The Percentage of Time Consumed by Each Module (\%)}
	\begin{tabular}{cccccc}
		\toprule
		\textbf{Sequence } & \textbf{ODD} & \textbf{DI} & \textbf{CE-PGO} & \textbf{Total (s)} & {\textbf{Remark}} \\  
		\midrule
		NCLT-01-10  & 0.04 & 0.26 & 1.02 & 780 & Single Event. \\
		UTBM-04-18  & 0.06    & 0.15    & 0.95  & 1050 & Single Event. \\
		Campus-1   & 0.01     & 0.38     & 0.51        & 394  & Single Event. \\
		Campus-4   & 0.02     & 0.53     & 1.08        & 469  & Two Events. \\
		Campus-5   & 0.01     & 0.92     & 1.43        & 491  & Three Events. \\
		\bottomrule
	\end{tabular}
	\vspace{1pt}
	\begin{adjustwidth}{0cm}{0cm}
		\footnotesize{ ``ODD'' denotes over-degeneracy detection module, ``DI'' denotes dynamic initialization module, ``CE-PGO'' denotes constraint-enhanced pose graph optimization module.}
	\end{adjustwidth}
	\label{tab5}
\end{table}

\section{CONCLUSION}
In this paper, we present and validate the multi-map LiDAR-inertial system. During periods of over-degeneracy, this system significantly reduces localization and mapping errors and prevents system failure, thanks to innovative aspects such as the reliable over-degeneracy detection method, robust multi-map strategy and constraint-enhanced map fusion strategy. The system's multi-map management module and constraint-enhanced pose graph optimization back-end are independent of the LIO system, enabling seamless transplantation to any superior LIO framework.However, our work still has some limitations. The accuracy of map fusion highly depends on the precision of point cloud recognition and registration. If there is minimal overlap in point clouds, the accuracy of map fusion will be significantly reduced. In the future, we will consider integrating semantic information to improve the accuracy of point cloud recognition and registration, making the system more robust.


\bibliographystyle{IEEEtran}
\bibliography{ref}

\vfill

\end{document}